\documentclass[11pt,a4paper, table, xcdraw]{article}
\usepackage[hyperref]{emnlp2020}
\usepackage{times}
\usepackage{graphicx} 
\usepackage{subfig}
\usepackage{booktabs}
\usepackage{amsmath}
\usepackage{latexsym}

\usepackage{microtype}
\usepackage{booktabs}

\usepackage{cleveref}
\crefname{section}{\S}{\S\S}
\Crefname{section}{\S}{\S\S}
\crefname{table}{Table}{}
\crefname{figure}{Fig.}{}
\crefname{algorithm}{alg.}{}
\crefname{equation}{eq.}{}
\crefname{appendix}{App.}{}
\crefformat{section}{\S#2#1#3}  

\aclfinalcopy 
\def\aclpaperid{2792} 

\DeclareMathOperator*{\argmin}{argmin}

\title{Tired of Topic Models? Clusters of Pretrained Word Embeddings\\Make for Fast and Good Topics too!}

\author{Suzanna Sia \hspace*{2em} Ayush Dalmia \hspace*{2em} Sabrina J. Mielke\\
  Department of Computer Science \\
  Johns Hopkins University  \\
  Baltimore, MD, USA \\
  \texttt{ssia1@jhu.edu}, \texttt{adalmia1@jhu.edu}, \texttt{sjmielke@jhu.edu} \\}

\date{}

\begin{document}
\maketitle
\begin{abstract}
Topic models are a useful analysis tool to uncover the underlying themes within document collections. The dominant approach is to use probabilistic topic models that posit a generative story, but in this paper we propose an alternative way to obtain topics: clustering pre-trained word embeddings while incorporating document information for weighted clustering and reranking top words. We provide benchmarks for the combination of different word embeddings and clustering algorithms, and analyse their performance under dimensionality reduction with PCA. The best performing combination for our approach performs as well as classical topic models, but with lower runtime and computational complexity.
\end{abstract}

\section{Introduction}

Topic models are the standard approach for exploratory document analysis \cite{boyd2017applications}, which aims to uncover main themes and underlying narratives within a corpus. But in times of distributed and even contextualized embeddings, are they the only option?

This work explores an alternative to topic modeling by casting `key themes' or `topics' as \emph{clusters of word types} under the modern distributed representation learning paradigm: unsupervised pre-trained word embeddings provide a representation for each word type as a vector, allowing us to cluster them based on their distance in high-dimensional space. The goal of this work is not to strictly outperform, but rather to benchmark
standard clustering of modern embedding methods against the classical approach of Latent Dirichlet Allocation \citep[LDA;][]{blei2003latent}.

We restrict our study to influential embedding methods and focus on centroid-based clustering algorithms as they provide a natural way to obtain the top words in each cluster based on distance from the cluster center.\footnote{We found that using non-centroid-based hierarchical, or density based clustering algorithms like DBScan resulted in worse performance and more hyperparameters to tune.}

Aside from reporting the best performing combination of word embeddings and clustering algorithm, we are also interested in whether there are consistent patterns: embeddings which perform consistently well across clustering algorithms might be good representations for unsupervised document analysis, clustering algorithms that perform consistently well are more likely to generalize to future word embedding methods.

To make our approach reliably work as well as LDA, we incorporate corpus frequency statistics directly into the clustering algorithm, and quantify the effects of two key methods, 1) weighting terms during clustering and 2) reranking terms for obtaining the top $J$ representative words. Our contributions are as follows:

\begin{itemize}
\item We systematically apply centroid-based clustering algorithms on top of a variety of pre-trained word embeddings and embedding methods for document analysis.

\item Through weighted clustering and reranking of top words we obtain sensible topics; the best performing combination is comparable with LDA, but with smaller time complexity and empirical runtime.

\item We show that further speedups are possible by reducing the embedding dimensions by up to 80\% using PCA.

\end{itemize}

\section{Related Work and Background}

Analyzing documents by clustering word embeddings is a natural idea---clustering has been used for readability assessment \cite{cha2017language}, argument mining \cite{reimers2019classification}, document classification and document clustering \cite{sano2017distributed}, \emph{inter alia}. So far, however, clustering word embeddings has not seen much success for the purposes of topic modeling. While many modern efforts have attempted to \emph{incorporate} word embeddings \emph{into} the probabilistic LDA framework \cite{liu2015topical, nguyen2015improving, das2015gaussian,zhao2017word,  batmanghelich2016nonparametric, xun2017correlated, dieng2019topic}, relatively little work has examined the feasibility of \emph{clustering embeddings directly}.

\citet{xie2013integrating} and \citet{viegas2019cluwords} first cluster documents and subsequently find words within each cluster for document analysis. \citet{sridhar2015unsupervised} targets short texts where LDA performs poorly in particular, fitting GMMs to learned word2vec representations. \Citet{de2019detecting} cluster using self-organising maps, but provide only qualitative results.

In contrast, our proposed approach is straightforward to implement, feasible for regular length documents, requires no retraining of embeddings, and yields qualitatively and quantitatively convincing results.
We focus on centroid based k-means (KM), Spherical k-means (SK), and k-medoids (KD) for hard clustering, and von Mises-Fisher Models (VMFM) and Gaussian Mixture Models (GMM) for soft clustering;
as pre-trained embeddings we consider word2vec \cite{mikolov2013efficient}, GloVe \cite{jeffreypennington2014glove}, FastText \cite{bojanowski2017enriching}, Spherical \cite{meng2019spherical}, ELMo \cite{Peters:2018}, and BERT \cite{devlin2018bert}.

\section{Methodology}

After preprocessing and extracting the vocabulary from our training documents, each word type is converted to its embedding representation (averaging all of its tokens for contextualized embeddings; details in \cref{sec:preprocess}).
Following this we apply the various clustering algorithms on the entire training corpus vocabulary to obtain $k$ clusters, using weighted (\cref{sec:weighting}) or unweighted word types. After the clustering algorithm has converged, we obtain the top $J$ words (\cref{sec:get_top_k}) from each cluster for evaluation. Note that one potential shortcoming of our approach is the possibility of outliers forming their own cluster, which we leave to future work.

\subsection{Obtaining top-J words} \label{sec:get_top_k}
In traditional topic modeling (LDA), the top $J$ words are those with highest probability under each topic-word distribution. For centroid based clustering algorithms, the top words of some cluster $i$ are naturally those closest to the cluster center $c^{(i)}$, or with highest probability under the cluster parameters.
Formally, this means choosing the set of types $J$ as
\[
    \argmin\limits_{J\colon |J|=10} \sum_{j \in J}
    \begin{cases}
        \|c^{(i)}-x_j\|_2^2 & \text{for KM/KD}, \\
        \cos(c^{(i)}, x_j) & \text{for SK}, \\  
        f(x_j \mid c^{(i)}, \Sigma_i) & \text{for GMM/VMFM}.\\
    \end{cases}
\]
\mbox{}\\[.5em]
Our results in \cref{sec:results} focus on KM and GMM, as we observe that k-medoids, spherical KM and von Mises-Fisher tend to perform worse than KM and GMM (see \cref{sec:kmedoids}, \cref{sec:spk}).\\[-1em]

Note that it is possible to extend this approach to obtain the top \emph{topics given a document}: compute similarity scores between learned topic cluster centers and all word embeddings from that particular document, and normalize them using softmax to obtain a (non-calibrated) probability distribution.\\[-1em]

Crucial to our method is the incorporation of corpus statistics on top of vanilla clustering algorithms, which we will describe in the remainder of this section.

\subsection{Weighting while clustering}\label{sec:weighting}

\begin{figure}[tb]
    \centering
    \includegraphics[width=1\linewidth]{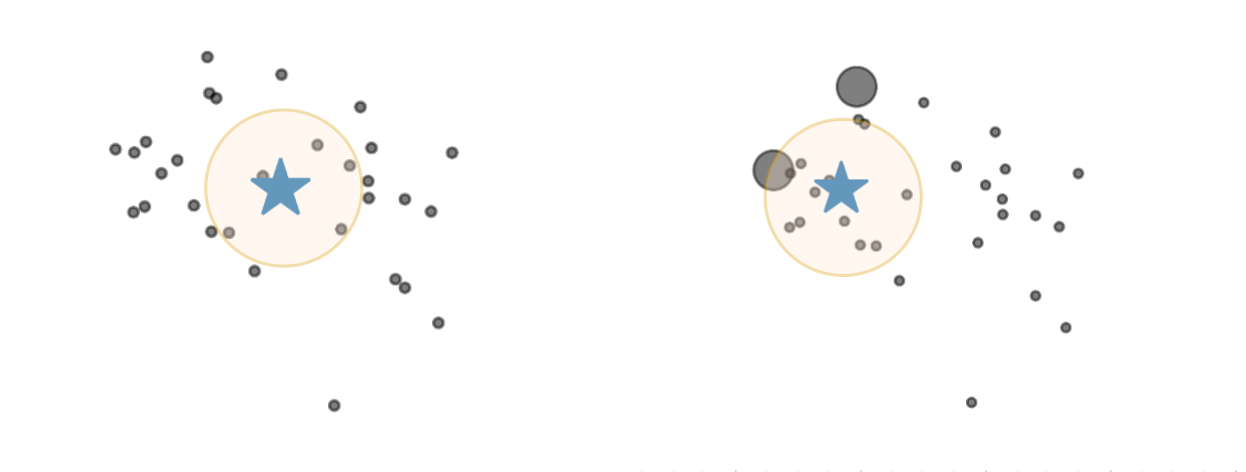}\\[-0.3em]
    \caption{The figure on the left shows the cluster center ($\star$) without weighting, while the figure on the right shows that after weighting (larger points have higher weight) a hopefully more representative cluster center is found. Note that top words based on distance from the cluster center could still very well be low frequency word types, motivating reranking (\cref{sec:reranking}).}
    \label{fig:cluster_ex}
\end{figure}

The intuition of \emph{weighted clustering} is based on the formulation of classical LDA which models the probability of the word type $t$ belonging to a topic $i$ as $\frac{N_{t,i} + \beta_t }{\sum_{t'} N_{t'i}+\beta_{t'}}$, where $N_{t,i}$ refers to the number of times word type $t$ has been assigned to topic $i$, and $\beta$ is a parameter of the Dirichlet prior on the per-topic word distribution. In our case, illustrated by the schematic in \cref{fig:cluster_ex}, weighting is a natural way to account for the frequency effects of vocabulary terms during clustering.

\subsection{Reranking when obtaining topics}\label{sec:reranking}
When obtaining the top-$J$ words that make up a cluster's topic, we also consider \emph{reranking} terms, as there is no guarantee that words closest to cluster centers are important word types. We will show in \cref{tab:words} that without reranking, clustering yields ``sensible'' topics but low NPMI scores.

\subsection{Which corpus statistics?}
To incorporate corpus statistics into the clustering algorithm, we examine three different schemes\footnote{We also experimented with various scaling methods such as robust scaling, logistic-sigmoid, and log transform but found that these do not improve performance.} to assign weights to word types, where $n_{t}$ is the count of word type $t$ in corpus $D$, and $d$ is a document:
\begin{align}
    \text{\bf tf} &=\dfrac{n_t}{\sum_{t'} n_{t'}}\\
    \text{\bf tf-df} &=\text{tf} \cdot \dfrac{|\{d\in D \mid t\in d\}|}{|D|}\\
    \text{\bf tf-idf} &= \text{tf} \cdot \log\left(\dfrac{|D|}{|\{d\in D \mid t\in d\}| + 1}\right)
\end{align}

These scores can now be used for \emph{weighting} word types when clustering (${\color{gray}\diamond}^w$), \emph{reranking} top 100 words (${\color{gray}\diamond}_r$) after, both (${\color{gray}\diamond}^w_r$), or neither (simply ${\color{gray}\diamond}$). We find that simply using \textbf{tf} outperforms the other weighting schemes (\cref{sec:reranking-schemes}). Our results and subsequent analysis in \cref{sec:results} uses \textbf{tf} for weighting and reranking.

\section{Computational Complexity} \label{sec:complexity}
The complexity of KM is $O(t k n m)$, and of GMM is $O(t k n m^3)$, for $t$ iterations,\footnote{In general, $t$ required for convergence differs for clustering algorithm and embedding representation. However we can specify the maximum number of iterations as a constant factor for worst case analysis.} $k$ clusters (topics), $n$ word types (unique vocabulary), and $m$ embedding dimensions. Weighted variants have a one-off cost of weight initialization, and contribute a constant factor when recalulculating the centroid during clustering. Reranking has an additional $O(n\cdot \log (n_k))$ factor, where $n_k$ is the average number of elements in a cluster. In contrast, LDA via collapsed Gibbs sampling has a complexity of $O(tkN)$, where $N$ is the number of all tokens, so when $N \gg n$, clustering methods can potentially achieve better performance-complexity tradeoffs.

Note that running ELMo and BERT over documents also requires iterating over all tokens, but only once, and not for every topic and iteration.

\subsection{Cost of obtaining Embeddings}

For readily available pretrained word embeddings such as word2vec, FastText, GloVe and Spherical, the embeddings can be considered as `given' as the practioner does not need to generate these embeddings from scratch. However for contextual embeddings such as ELMo and BERT, there is additional computational cost in obtaining these embeddings before clustering, which requires passing through RNN and transformer layers respectively. This can be trivially parallelised by batching the context window (usually a sentence). We use standard pretrained ELMo and BERT models in our experiments and therefore do not consider the runtime of training these models from scratch.

\section{Experimental Setup}\label{sec:experiments}

\begin{table*}[tb]
\resizebox{\textwidth}{!}{\begin{tabular}{r||rr|rr|rr|rr||rr|rr|rr|rr}
 & \multicolumn{8}{c||}{\em Reuters} & \multicolumn{8}{c}{\em 20 Newsgroups} \\
 & \multicolumn{2}{c|}{${\color{gray}\diamond}$} & \multicolumn{2}{c|}{${\color{gray}\diamond}^w$} & \multicolumn{2}{c|}{${\color{gray}\diamond}_r$} & \multicolumn{2}{c||}{${\color{gray}\diamond}^w_r$} & \multicolumn{2}{c|}{${\color{gray}\diamond}$} & \multicolumn{2}{c|}{${\color{gray}\diamond}^w$} & \multicolumn{2}{c|}{${\color{gray}\diamond}_r$} & \multicolumn{2}{c}{${\color{gray}\diamond}^w_r$} \\
 & KM & GMM & KM & GMM & KM & GMM & KM & GMM & KM & GMM & KM & GMM & KM & GMM & KM & GMM \\
 \toprule
Word2vec & \cellcolor{green!0} -0.39 & \cellcolor{green!0} -0.47 & \cellcolor{green!7} -0.21 & \cellcolor{green!18} -0.09 & \cellcolor{green!38} 0.02 & \cellcolor{green!36} 0.01 & \cellcolor{green!40} 0.03 & \cellcolor{green!54} 0.08 & \cellcolor{green!0} -0.21 & \cellcolor{green!3} -0.10 & \cellcolor{green!2} -0.11 & \cellcolor{green!32} 0.13 & \cellcolor{green!47} 0.18 & \cellcolor{green!40} 0.16 & \cellcolor{green!50} 0.19 & \cellcolor{green!53} 0.20 \\
ELMo & -0.73 & \cellcolor{green!0} -0.55 & \cellcolor{green!0} -0.43 & \cellcolor{green!34} 0.00 & \cellcolor{green!17} -0.10 & \cellcolor{green!20} -0.08 & \cellcolor{green!30} -0.02 & \cellcolor{green!48} 0.06 & -0.56 & \cellcolor{green!2} -0.13 & \cellcolor{green!0} -0.38 & \cellcolor{green!47} 0.18 & \cellcolor{green!32} 0.13 & \cellcolor{green!35} 0.14 & \cellcolor{green!40} 0.16 & \cellcolor{green!50} 0.19 \\
GloVe & -0.67 & \cellcolor{green!0} -0.59 & \cellcolor{green!26} -0.04 & \cellcolor{green!36} 0.01 & \cellcolor{green!3} -0.27 & \cellcolor{green!28} -0.03 & \cellcolor{green!36} 0.01 & \cellcolor{green!45} 0.05 & \cellcolor{green!0} -0.18 & \cellcolor{green!2} -0.12 & \cellcolor{green!18} 0.06 & \cellcolor{green!69} 0.24 & \cellcolor{green!61} 0.22 & \cellcolor{green!65} 0.23 & \cellcolor{green!65} 0.23 & \cellcolor{green!65} 0.23 \\
Fasttext & -0.68 & -0.70 & \cellcolor{green!0} -0.46 & \cellcolor{green!20} -0.08 & \cellcolor{green!34} 0.00 & \cellcolor{green!34} 0.00 & \cellcolor{green!48} 0.06 & \cellcolor{green!63} 0.11 & \cellcolor{green!0} -0.32 & \cellcolor{green!0} -0.20 & \cellcolor{green!0} -0.18 & \cellcolor{green!57} 0.21 & \cellcolor{green!69} 0.24 & \cellcolor{green!65} 0.23 & \cellcolor{green!74} 0.25 & \cellcolor{green!69} 0.24 \\
Spherical & \cellcolor{green!0} -0.53 & -0.65 & \cellcolor{green!21} -0.07 & \cellcolor{green!57} 0.09 & \cellcolor{green!36} 0.01 & \cellcolor{green!24} -0.05 & \cellcolor{green!60} 0.10 & \cellcolor{green!67} 0.12 & \cellcolor{green!6} -0.05 & \cellcolor{green!0} -0.24 & \cellcolor{green!69} 0.24 & \cellcolor{green!65} 0.23 & \cellcolor{green!74} 0.25 & \cellcolor{green!61} 0.22 & \cellcolor{green!79} 0.26 & \cellcolor{green!69} 0.24 \\
BERT & \cellcolor{green!0} -0.43 & \cellcolor{green!8} -0.19 & \cellcolor{green!21} -0.07 & \cellcolor{green!67} 0.12 & \cellcolor{green!34} 0.00 & \cellcolor{green!32} -0.01 & \cellcolor{green!67} 0.12 & \cellcolor{green!78} 0.15 & \cellcolor{green!15} 0.04 & \cellcolor{green!35} 0.14 & \cellcolor{green!74} 0.25 & \cellcolor{green!74} 0.25 & \cellcolor{green!43} 0.17 & \cellcolor{green!50} 0.19 & \cellcolor{green!74} 0.25 & \cellcolor{green!74} 0.25 \\
\midrule
average & -0.57 & -0.52 & -0.21 & 0.01 & -0.06 & -0.03 & 0.05 & 0.10 & -0.21 & -0.11 & -0.02 & 0.21 & 0.20 & 0.20 & 0.23 & 0.23 \\
std. dev. & 0.14 & 0.18 & 0.19 & 0.09 & 0.12 & 0.03 & 0.05 & 0.04 & 0.21 & 0.13 & 0.25 & 0.05 & 0.04 & 0.04 & 0.04 & 0.02 \\
\bottomrule
\end{tabular}}
\caption{NPMI Results (higher is better) for pre-trained word embeddings and k-means (KM), and Gaussian Mixture Models (GMM). ${\color{gray}\diamond}^w$ indicates weighted and ${\color{gray}\diamond}_r$ indicates reranking of top words. For Reuters (left table), LDA has an NPMI score of 0.12, while GMM$^w_r$ BERT achieves 0.15. For 20NG (right), both LDA and KM$^w_r$ Spherical achieve a score of 0.26. All results are averaged across 5 random seeds.}
\label{tab:results}
\end{table*}

Our implementation is freely available online.\footnote{\url{https://github.com/adalmia96/Cluster-Analysis}}

\subsection{Datasets}
We use the 20 newsgroup dataset (20NG) which contains around $18000$ documents and $20$ categories,\footnote{\url{http://qwone.com/~jason/20Newsgroups/}} and a subset of Reuters21578\footnote{\url{https://www.nltk.org/book/ch02.html}} which contains around $10000$ documents.

\subsection{Evaluation (Topic Coherence)}

We adopt a standard 60-40 train-test split for 20NG and 70-30 for Reuters.

The top $10$ words (\cref{sec:get_top_k}) were evaluated using \emph{normalized pointwise mutual information} \cite[NPMI;][]{bouma2009normalized} which has been shown to correlate with human judgements \cite{lau2014machine}. NPMI ranges from $[-1,1]$ with $1$ indicating perfect association. The train split is used to obtain the top topic words in an unsupervised fashion (we do not use any document labels), and the test split is used to evaluate the ``topic coherence'' of these top words. NPMI scores are averaged across all topics.

For both datasets we use $20$ topics; which gives best NPMI out of $20$, $50$, $100$ topics for Reuters, and is the ground truth number for 20NG. The NPMI scores presented in \cref{tab:results} are averaged across cluster centers initialized using 5 random seeds.

\subsection{Preprocessing}
\label{sec:preprocess}

We lowercase tokens, remove stopwords, punctuation and digits, and exclude words that appear in less than 5 documents and appear in long sentences of more than 50 words, removing email artifacts and noisy token sequences which are not valid sentences. An analysis on the effect of rare word removal can be found in \cref{sec:results-weighting}.

For contextualized word embeddings (BERT and ELMo), sentences served as the context window to obtain the token representations. Subword representations were averaged for BERT, which performs better than just using the first subword.

\section{Results and Discussion} \label{sec:results}

Our main results are shown in \cref{tab:results}.

\begin{figure*}%
    \centering
    \includegraphics[width=\linewidth]{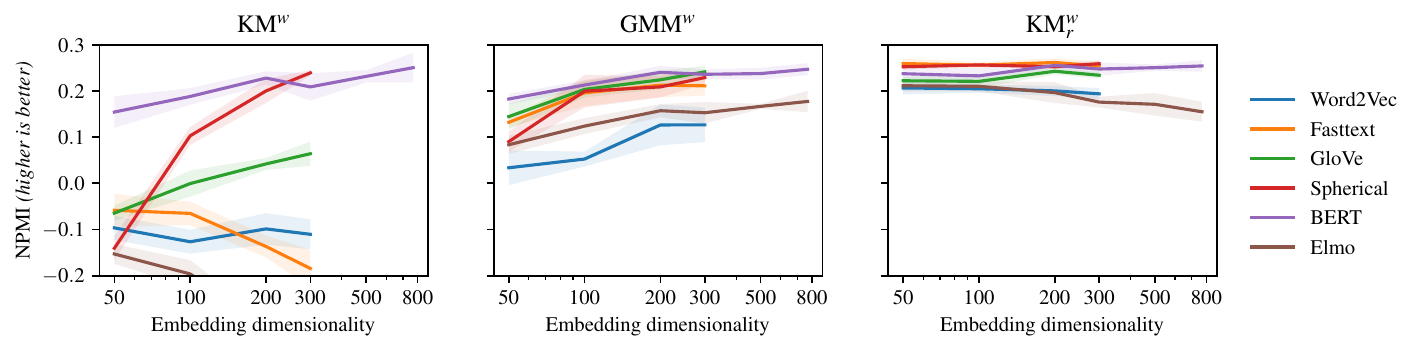}
    \caption{ Plots showing the effect of PCA dimension reduction on different embedding and clustering algorithms. KM$^w_r$ which we advocate over GMMs for efficiency, allows for dimension reduction of up to 80\%.}
    \label{fig:dim_reduce}%
\end{figure*}

\subsection{Runtime}
Running LDA with \texttt{MALLET} \citep{McCallumMALLET} takes a minute, but performs no better than KM$^w_r$, which takes little more than 10 seconds on CPU using \texttt{sklearn} \citep{scikit-learn}, and 3-4 seconds using a simple implementation using JAX \citep{jax2018github} on GPU.

\subsection{Weighting}\label{sec:results-weighting}
\begin{figure}[t!]
    \centering
    \includegraphics[width=\linewidth]{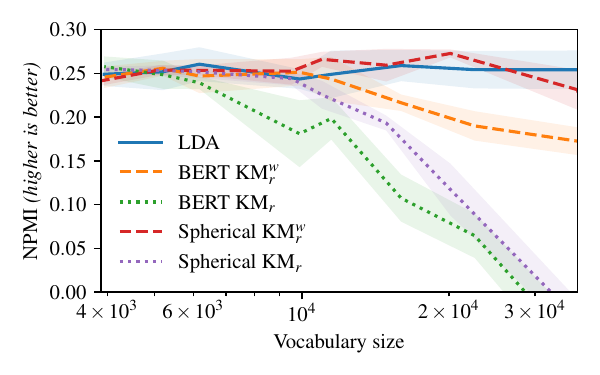}\\[-0.3em]
    \caption{NPMI as a function of vocabulary size reduced by term frequency on 20NG. Embeddings are more sensitive to noisy vocabulary (infrequent terms) than LDA, but reweighting (${\color{gray}\diamond}^w$) helps to alleviate this.}
    \label{fig:vocabsize}
\end{figure}

From \cref{tab:results}, we see that reranking and weighting greatly improves clustering performance across different embeddings. As a first step to uncover why, we investigate how sensitive our methods are to restricting the clustering to only \emph{frequently appearing word types}. Visualized in \cref{fig:vocabsize}, we find that as we vary the cutoff term frequency, thus changing the vocabulary size and allowing more rare words on the x-axis, NPMI is more affected for the models without reweighting. This suggests that reweighting using term frequency is effective for clustering without the need for ad-hoc restriction of infrequent terms---without it, all combinations perform poorly compared to LDA. In general, GMM outperforms KM for both weighted and unweighted variants averaged across all embedding methods ($p<0.05$).\footnote{Two-tailed t-test for GMM$^w$ vs KM$^w$.}

\subsection{Reranking}
For KM, extracted topics before reranking results in reasonable looking themes, but scores poorly on NPMI. Reranking strongly improves KM on average ($p<0.02$) for both Reuters and 20NG. Examples before and after reranking are provided in \cref{tab:words}. This indicates that while cluster centers are centered around valid themes, they are surrounded by low frequency word types.

\begin{table}[t!]
\resizebox{\columnwidth}{!}{\begin{tabular}{ll|ll}

    \multicolumn{2}{l|}{BERT (topic12)} & \multicolumn{2}{l}{Spherical (topic19)} \\
 \toprule
  KM & KM$_r$ &  KM & KM$_r$ \\
 \midrule
  vram & drive & detector & earth \\
  vesa & hard & electromagnetic & nasa \\
  cmos & card & magnetic & satellite \\
  portable & computer & spectrometer & orbit \\
  micron & chip & infrared & surface \\
  nubus & machine & optical & energy \\
  digital & video & velocity & radar \\
  machine & hardware & radiation & solar \\
  motherboards & clipper & solar & spacecraft \\
  hardware & controller & telescope & electrical \\
 \bottomrule
 \textit{\footnotesize\color{gray} NPMI:} -0.36 & \textit{\footnotesize\color{gray} NPMI:} \textbf{0.15} & \textit{\footnotesize\color{gray} NPMI:} -0.01 & \textit{\footnotesize\color{gray} NPMI:} \textbf{0.36}

\end{tabular}}

\caption{Top 10 words in a topic on 20NG and overall NPMI, for k-means (KM) before and after reranking (KM$_r$): reranking clearly improves NPMI for BERT and Spherical.}

\label{tab:words}
\end{table}

We observe that when applying reranking to GMM$^w$ the gains are much less pronounced than KM$^w$. The top topic words before and after reranking for BERT-GMM$^w$ have an average Jaccard similarity score of 0.910, indicating that the cluster centers learned by weighted GMMs are already centered at word types of high frequency in the training corpus.

\subsection{Embeddings}
Spherical embeddings and BERT perform consistently well across both datasets. For 20NG, KM$^w_r$ Spherical and LDA both achieve $0.26$ NPMI.
For Reuters, GMM$^w_r$ BERT achieves the top NPMI score of $0.15$ compared to $0.12$ of LDA.
Word2vec and ELMo (using only the last layer\footnote{Selected as best performing by manually testing 13 different mixing ratios.}) perform poorly compared to the other embeddings.
FastText and GloVe can achieve similar performance to BERT on 20NG but are slightly inferior on Reuters.

Training or fine-tuning embeddings on the given data prior to clustering could potentially achieve better performance, but we leave this to future work.

\subsection{Qualitative results}
We find that our approach yields a greater diversity within topics as compared to LDA while achieving comparable coherence scores (\cref{sec:qualitative}). Such topics are arguably more valuable for exploratory analysis.

\subsection{Dimensionality Reduction}
We apply PCA to the word embeddings before clustering to investigate the amount of redundancy in the dimensions of large embeddings, which impact clustering complexity (\cref{sec:complexity}).
With reranking, the dimensions of all embeddings can be reduced by more than 80\% (\cref{fig:dim_reduce}). 

We observe that KM$^w_r$ can consistently reduce the number of dimensions across different embedding types without loss of performance. Although GMM$^w$ does not require reranking for good performance, it's cubic complexity indicates that KM$^w_r$ might be preferred in practical settings.

\section{Conclusion}
We outlined a methodology for clustering word embeddings for unsupervised document analysis, and presented a systematic comparison of various influential embedding methods and clustering algorithms. Our experiments suggest that pre-trained word embeddings (both contextualized and non-contextualized), combined with tf-weighted k-means and tf-based reranking, provide a viable alternative to traditional topic modeling at lower complexity and runtime.

\section*{Acknowledgments}
We thank Aaron Mueller, Pamela Shapiro, Li Ke, Adam Poliak, Kevin Duh and the anonymous reviewers for their feedback.

\bibliography{anthology,clusterembeds}
\bibliographystyle{acl_natbib}

\newpage
\appendix

\section{k-means (KM) vs k-medoids (KD)}\label{sec:kmedoids}

To further understand the effect of other centroid based algorithms on topic coherence, we also applied the k-medoids (KD) clustering algorithm.  KD is a hard clustering algorithm similar to KM but less sensitive to outliers.

\begin{table}[b]
\centering
\begin{tabular}{l|rr|rr|}
                     & KM       & KD     & KM$_r$ & KD$_r$    \\
                      \hline
Word2Vec             & -0.21 &	-0.32 &	0.18 & 0.12  \\
FastText             & -0.33 &	-0.39 &	0.24 &	0.19       \\
GloVe                & -0.18& -0.43 &	0.22 & 0.08    \\
BERT                 & 0.04 &	-0.06 &  0.17 &	0.15    \\
ELMo               & -0.56 &	-0.56 &	0.13 &	0.12    \\
Spherical            & -0.05&	-0.07 &	0.25 & 0.22       \\ \hline
average & -0.22	& -0.31 &	0.20 &	0.15 \\
std. dev. & 0.21 &	0.20 &	0.04 &	0.05
\end{tabular}
   \caption{Results for pre-trained word embeddings and k-means (KM) and k-medoids (KD). $r$ indicates reranking of top words using term frequency.}
    \label{tab:kmedoids}
\end{table}

As we can see in \cref{tab:kmedoids}, in all cases KD usually did as well or worse than KM. KD also did relatively poorly after frequency reranking. Where KD did do better than KM, the difference is not very striking and the NPMI scores were still quite below the other top performing models.

\section{Results for Spherical k-means and Von Mises-Fisher Mixture}\label{sec:spk}

\Cref{tab:resultsspk} shows the overall bad performance of spherical clustering methods, specifically Spherical k-Means (SKM) and von-Mises-Fisher mixtures (VMFM).

\begin{table*}[bt]
\centering
\begin{tabular}{r||rr|rr|rr|rr}
 & \multicolumn{8}{c}{\em Reuters} \\
 & \multicolumn{2}{c|}{${\color{gray}\diamond}$} & \multicolumn{2}{c|}{${\color{gray}\diamond}^w$} & \multicolumn{2}{c|}{${\color{gray}\diamond}_r$} & \multicolumn{2}{c}{${\color{gray}\diamond}^w_r$} \\
 & SKM & VMFM & SKM & VMFM & SKM & VMFM & SKM & VMFM \\
 \toprule
Word2vec  & -0.70 & -0.85 & -0.43 & -0.88 & -0.16 & -0.05 & -0.19 & -0.05 \\
ELMo      & -0.74 & -0.88 & -0.37 & -0.87 & -0.14 & -0.10 &  0.00 & -0.12 \\
GloVe     & -0.52 & -0.88 & -0.11 & -0.88 &  0.00 & -0.18 &  0.06 & -0.17 \\
Fasttext  & -0.85 & -0.89 & -0.65 & -0.87 & -0.18 & -0.08 & -0.18 & -0.10 \\
Spherical & -0.50 & -0.81 & -0.08 & -0.82 &  0.01 & -0.07 &  0.10 & -0.09 \\
BERT      & -0.40 & -0.88 & -0.06 & -0.65 & -0.03 & -0.14 &  0.11 & -0.16 \\
average   & -0.62 & -0.87 & -0.28 & -0.83 & -0.08 & -0.10 & -0.02 & -0.12 \\
std. dev. &  0.17 &  0.03 &  0.24 &  0.09 &  0.09 &  0.05 &  0.14 &  0.04 \\
\end{tabular}

\mbox{}\\[1em]

\begin{tabular}{r||rr|rr|rr|rr}
 & \multicolumn{8}{c}{\em 20 Newsgroups} \\
 & \multicolumn{2}{c|}{${\color{gray}\diamond}$} & \multicolumn{2}{c|}{${\color{gray}\diamond}^w$} & \multicolumn{2}{c|}{${\color{gray}\diamond}_r$} & \multicolumn{2}{c}{${\color{gray}\diamond}^w_r$} \\
 & SKM & VMFM & SKM & VMFM & SKM & VMFM & SKM & VMFM \\
 \toprule
Word2vec  & -0.37 & -0.59 & -0.17 & -0.88 &  0.15 &  0.17 &  0.14 &  0.16 \\
ELMo      & -0.52 & -0.66 & -0.30 & -0.87 &  0.16 &  0.10 &  0.20 &  0.12 \\
GloVe     &  0.00 & -0.62 &  0.23 & -0.88 &  0.25 &  0.13 &  0.24 &  0.14 \\
Fasttext  & -0.60 & -0.58 & -0.26 & -0.54 &  0.12 &  0.19 &  0.14 &  0.19 \\
Spherical & -0.04 & -0.54 &  0.22 & -0.82 &  0.25 &  0.22 &  0.25 &  0.21 \\
BERT      &  0.06 & -0.62 &  0.22 & -0.65 &  0.23 &  0.11 &  0.25 &  0.10 \\
average   & -0.24 & -0.60 & -0.01 & -0.77 &  0.19 &  0.15 &  0.20 &  0.15 \\
std. dev. &  0.29 &  0.04 &  0.26 &  0.14 &  0.06 &  0.05 &  0.05 &  0.04 \\
\end{tabular}
\caption{NPMI Results (higher is better) for pre-trained word embeddings and Spherical k-means (SKM), and von Mises-Fisher Mixtures (VMFM). ${\color{gray}\diamond}^w$ indicates weighted and ${\color{gray}\diamond}_r$ indicates reranking of top words. }
\label{tab:resultsspk}
\end{table*}

\begin{table}[bt]
\centering
\begin{tabular}{l|rrrr}
                     & TF  & TF-IDF   & TF-DF \\ \hline
Word2Vec             & 0.18 &	0.15 &	0.17  \\
FastText             & 0.24 & 	0.23 &	0.23  \\
GloVe                & 0.22 &	0.17 &	0.21  \\
BERT                 & 0.17 &	0.15 &	0.17  \\
ELMo                & 0.13 &	0.09 &	0.14  \\
Spherical            &0.25 &	0.22 &	0.24  \\ \hline
average             & 0.20	& 0.17 &	0.19  \\
std. dev.           & 0.04 & 0.05 & 	0.04 
\end{tabular}
   \caption{Results for k-means (without weighting) with pre-trained word embeddings using different reranking metrics : TF, TF-IDF, and TF-DF.  }
    \label{tab:tfwithoutrank}
\end{table}

\begin{table}[bt]
\centering
\begin{tabular}{l|rrr}
                      & ${\color{gray}\diamond}^w$ TF       & ${\color{gray}\diamond}^w$ TF-IDF    & ${\color{gray}\diamond}^w$ TF-DF     \\ \hline
Word2Vec              &0.19 &	0.17 &	0.20     \\
FastText              & 0.25 &	0.25 &	0.25       \\
GloVe                 & 0.23 &	0.21 &	0.23         \\
BERT                  & 0.25 &	0.24  &	0.25        \\
ELMo                  & 0.16 &	0.15 &	0.16     \\
Spherical             & 0.26 &	0.24 &	0.25        \\ \hline
average               & 0.23 &	0.21 &	0.22         \\
std. dev.            &  0.04 &	0.04 &	0.04
\end{tabular}
   \caption{Results for k-means (weighted) pre-trained word embeddings using different reranking metrics: TF, TF-IDF and TF-DF weighted with term frequency.}
    \label{tab:tfwithrank}
\end{table}

\section{Comparing Different Reranking Schemes}\label{sec:reranking-schemes}

We present the results for using different reranking schemes for KM (\cref{tab:tfwithoutrank}) and Weighted KM for Frequency (\cref{tab:tfwithrank}).

We can see that compared to the TF results in the main paper, other schemes for reranking such as aggregated TF-IDF and TF-DF improve over the original hard clustering, but fare worse in comparison with reranking with TF.

\section{Qualitative Comparison of Topics Generated}\label{sec:qualitative}
We present the different topics generated using LDA (\cref{tab:lda_reuters})  and topics generated using BERT KM$^w_r$ for the Reuters dataset (\cref{tab:bertkmnr_reuters}).
Note that unlike LDA, which uses the highest posterior probability allowing duplicate words to appear in duplicate topics, using a hard clustering algorithm for assignment mean that each word is assigned to one topic only.
We can see compared to the LDA topics which tend to contain topics mostly regarding wealth and profits, clustering with BERT KM$^w_r$ introduces new topics in involving locations and corporate positions. We see overall that using clustering allows for a discovery for a greater diversity of topics due to the greater diversity of words over all the topics.

\begin{table*}[p!]
\centering
\begin{tabular}{l|r} 
 Top 10 Word for Each Topic & NPMI \\
\hline
 dollar rate rates exchange currency market dealers central interest point & 0.369\\
year growth rise government economic economy expected domestic inflation report  & 0.355 \\
 gold reserves year tons company production exploration ounces feet mine  & 0.290 \\
 billion year rose dlrs fell marks earlier figures surplus rise & -0.005 \\
 year tonnes crop production week grain sugar estimated expected area  & 0.239 \\
  dlrs company sale agreement unit acquisition assets agreed subsidiary sell & -0.043\\
 bank billion banks money interest market funds credit debt loans & 0.239 \\
 tonnes wheat export sugar tonne exports sources shipment sales week &  0.218 \\
plan bill industry farm proposed government administration told proposal change &  0.212 \\
 prices production price crude output barrels barrel increase demand industry & 0.339 \\
 group company investment stake firm told companies capital chairman president & 0.191 \\
trade countries foreign officials told official world government imports agreement & 0.298 \\
 offer company shares share dlrs merger board stock tender shareholders & 0.074 \\
 shares stock share common dividend company split shareholders record outstanding & 0.277 \\
 dlrs year quarter earnings company share sales reported expects results & -0.037\\
 market analysts time added long analyst term noted high back & 0.316 \\
 coffee meeting stock producers prices export buffer quotas market price & 0.170  \\
 loss dlrs profit shrs includes year gain share mths excludes & -0.427\\
 spokesman today government strike union state yesterday workers officials told & 0.201 \\
 program corn dlrs prior futures price loan contract contracts cents & -0.287 \\
\end{tabular}
\caption{ NPMI Scores and Top 10 words for the topics generated using LDA for the Reuters dataset}
\label{tab:lda_reuters}
\end{table*}

\begin{table*}[p!]
\centering
\begin{tabular}{l|r} 
 Top 10 Word for Each Topic & NPMI \\
\hline
 rise increase growth fall change decline drop gains cuts rising & 0.238\\
 president chairman minister house baker administration secretary executive chief washington & 0.111\\
 make continue result include reduce open support work raise remain & 0.101\\
 january march february april december june september october july friday & 0.043\\
 year quarter week month earlier months years time period term & 0.146\\
 rose fell compared reported increased estimated revised adjusted unchanged raised & 0.196\\
 today major made announced recent full previously strong final additional & 0.125\\
 share stock shares dividend common cash stake shareholders outstanding preferred & 0.281\\
 dlrs billion tonnes marks francs barrels cents tonne barrel tons & -0.364\\
 sales earnings business operations companies products markets assets industries operating & 0.115\\
sale acquisition merger sell split sold owned purchase acquire held & 0.003 \\
board meeting report general commission annual bill committee association council & 0.106 \\
loss profit revs record note oper prior shrs gain includes & 0.221 \\
 company corp group unit firm management subsidiary trust pacific holdings & 0.058 \\
 prices price current total lower higher surplus system high average & 0.198\\
 offer agreement agreed talks tender plan terms program proposed issue & 0.138\\
 bank trade market rate exchange dollar foreign interest rates banks & 0.327\\
 told official added department analysts officials spokesman sources statement reuters & 0.181\\
 production export exports industry wheat sugar imports output crude domestic & 0.262\\
 japan government international world countries american japanese national states united & 0.251 \\
\end{tabular}
\caption{ NPMI Scores and Top 10 words for the topics generated using BERT KM$^w_r$   for the Reuters dataset}
\label{tab:bertkmnr_reuters}
\end{table*}

\end{document}